\pdfoutput=1

\documentclass[11pt]{article}

\usepackage[]{acl2023}

\usepackage{times}
\usepackage{latexsym}

\usepackage{inconsolata}
\usepackage{tcolorbox}
\usepackage{booktabs}
\usepackage{multirow}
\usepackage{arydshln}
\usepackage{float}
\usepackage{tabularx}
\usepackage{caption,subcaption}
\usepackage{colortbl}
\usepackage{amssymb}
\usepackage{pifont}

\usepackage{algorithm}
\usepackage{algpseudocode}

\usepackage[T1]{fontenc}

\definecolor{Gray}{gray}{0.92}
\definecolor{LightGray}{gray}{0.96}
\definecolor{LightCyan}{rgb}{0.92,0.968,0.968}
\definecolor{amber}{rgb}{1.0, 0.75, 0.0}%
\definecolor{neuesrot}{RGB}{207, 103, 102}%
\definecolor{lightblue}{RGB}{100, 181, 246}%
\definecolor{lightgreen}{RGB}{129, 199, 132}%

\newtheorem{finding}{Finding}
\usepackage{makecell}

\newenvironment{short_list}
  {\obeylines\everypar{\textbullet\hspace{\labelsep}}} 
  {} 

\usepackage[utf8]{inputenc}

\usepackage{microtype}

\newcommand{\rparagraph}[1]{\vspace{1.6mm}\noindent\textbf{#1.}}

\usepackage{inconsolata}

\title{Zero-Shot Fact-Checking with Semantic Triples and Knowledge Graphs}

\author{Zhangdie Yuan \and Andreas Vlachos \\
        Department of Computer Science and Technology \\ University of Cambridge \\
        \texttt{zy317,av308@cam.ac.uk}}

\usepackage{tikz}
\usetikzlibrary{positioning}
\usepackage{amssymb}

\begin{document}
\maketitle
\begin{abstract}

Despite progress in automated fact-checking, most systems require a significant amount of labeled training data, which is expensive. In this paper, we propose a novel zero-shot method, which instead of operating directly on the claim and evidence sentences, decomposes them into semantic triples augmented using external knowledge graphs, and uses large language models trained for natural language inference. 
This allows it to generalize to adversarial datasets and domains that supervised models require specific training data for.
Our empirical results show that our approach outperforms previous zero-shot approaches on FEVER, FEVER-Symmetric, FEVER 2.0, and Climate-FEVER, while being comparable or better than supervised models on the adversarial and the out-of-domain datasets.
\end{abstract}

\section{Introduction}

Fact-checking is the task of assessing the truthfulness of a claim, and is well-studied across multiple disciplines. Traditionally, journalists perform such a task manually, which is time-consuming. More recently, automated fact-checking systems have become of interest
due to  the explosion of (mis)information on social media~\cite{adair2017progress, hassan-et-al-2017-check-worthy-claims}. In the NLP community, fact-checking is typically defined as a task consisting of three stages: claim detection, evidence retrieval, and claim verification~\cite{guo2022survey}.
In particular, verdict prediction assumes the evidence is retrieved from sources such as Wikipedia or the web, and aims to predict the verdict of a claim given the retrieved evidence, often as a three-way classification task \citep{thorne2018fever}: \textsc{Supports}, \textsc{Refutes}, and \textsc{NEI} (\textsc{Not Enough Info}).

\begin{figure*}[htp]
    \centering
    \includegraphics[width=\textwidth]{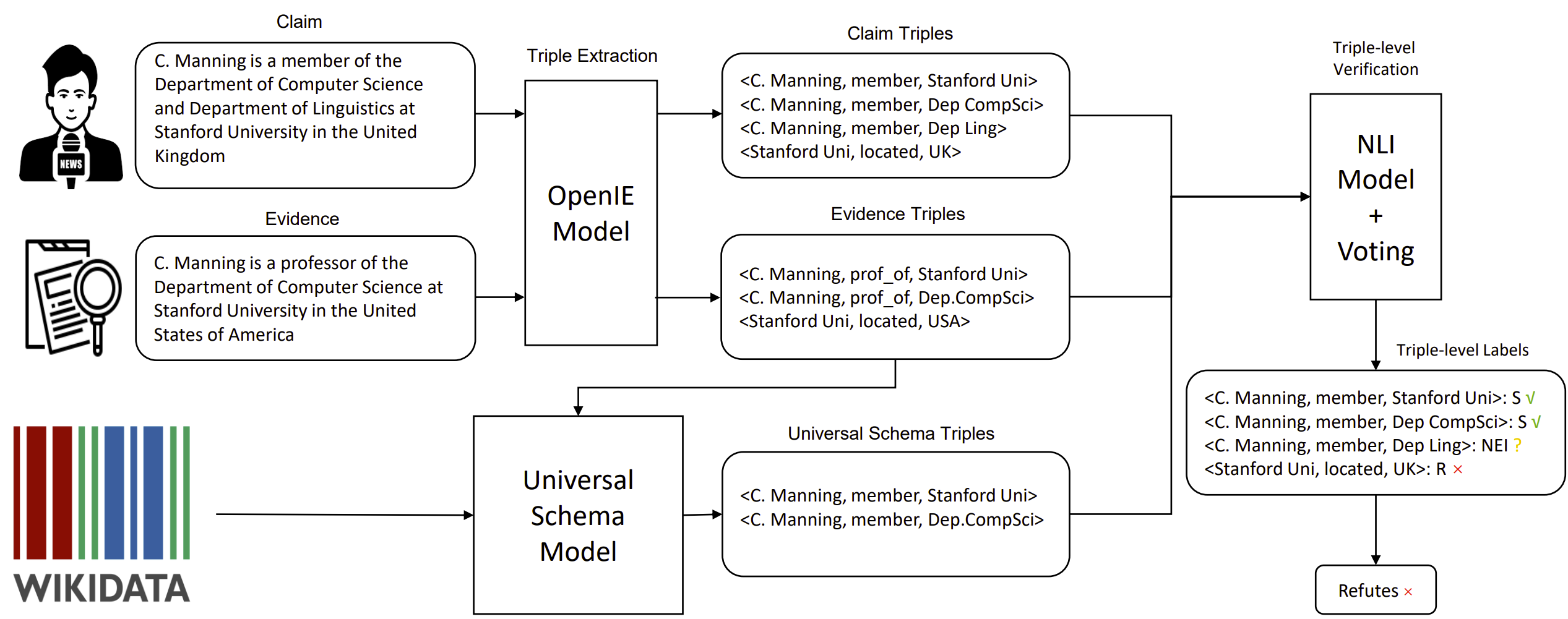}
    \caption{An overview of our zero-shot learning system. By harnessing Wikidata for training the universal schema model, incorporating on-demand training with evidence triples, and leveraging OpenIE for triple-level inference, our system achieves enhanced improvements. Label S stands for \textsc{Supports} and R stands for \textsc{Refutes}.}
    \vspace{-2mm}
    \label{fig:overview}
\end{figure*}

Recent work~\citep{dehaven2023bevers} has achieved strong results on canonical datasets like FEVER~\citep{thorne2018fever}, mostly relying on supervised approaches. However,  concerns have been expressed on whether these models learn language's and the task's nuances or merely leverage embedded biases and dataset idiosyncrasies. This argument~\citep{gururangan-etal-2018-annotation, poliak-etal-2018-hypothesis} gains empirical weight when such high-performing models are tested against adversarial fact-checking datasets such as FEVER-Symmetric~\citep{schuster-etal-2019-towards} and FEVER 2.0~\citep{thorne-etal-2019-fever2}. Their underperformance~\citep{Thorne19FEVER2} in these adversarial benchmarks exposes a lack of model robustness.

The narrative of this vulnerability extends to out-of-domain contexts as well. A pertinent example is the Climate-FEVER dataset—a platform for verifying real-world climate claims~\citep{diggelmann2020climate}. Supervised models, despite their commendable performance on the original FEVER dataset, suffer performance degradation when evaluated on Climate-FEVER. Additionally, earlier zero-shot fact-checking approaches~\citep{pan-etal-2021-zero, wright2022generating} hinge on synthetic data creation for training purposes. While this data emanates from factual evidence, it largely adheres to the domain boundaries of the originating dataset. Such inherent domain confinement curtails the model's capacity for broader generalization.

In this work we propose a zero-shot method utilizing semantic triples and knowledge graphs in conjunction with pretrained Natural Language Inference (NLI) models, and does not require training data for parameter learning.

In particular, we propose to extract triples from the claim and the evidence texts to form knowledge graphs and fill potential gaps in the evidence using a universal schema model~\citep{riedel-etal-2013-relation} on Wikidata and Wikipedia. Crucially, our method refrains from utilizing any annotated or synthetic training data, sidestepping the pitfalls of biases and dataset artifacts that can inadvertently be encoded into models. Additionally, by decomposing the original claim into triples, our method can harness the pre-trained NLI model's strengths more effectively. Both design choices position our approach to exhibit greater robustness when subjected to adversarial and out-of-domain evaluations.

As shown in Figure~\ref{fig:overview}, we follow a two-stage verification process: triple-level and claim-level. For triple-level, we employ NLI models pre-trained without further fine-tuning on any fact-checking training dataset, hence a zero-shot setting. For claim-level verification, we design a simple rule-based system relying on the triple verification. In Figure~\ref{fig:overview}, the process involves extracting claim and evidence sentences to generate triples. Subsequently, the universal schema is applied to obtain additional triples. The NLI model is then employed to assign triple-level labels, resulting in 2 \textsc{Supports}, 1 \textsc{Refutes}, and 1 \textsc{NEI}\footnote{In the context of our study, the NLI labels have been appropriately reconfigured to align with the FEVER labels.}. Finally, a rule-based system is utilized to derive claim-level verification. In this  example, since one claim triple is refuted, the entire claim is considered refuted. Note that we are able to use the ``gap'' triples filled by the universal schema model to retrieve better evidence. For example, \textit{<Manning, member\_of, Stanford>} is needed to verify the claim. However, such a triple is missing from the evidence triple extraction because the word \textit{member} is not mentioned in the evidence. Instead, \textit{<Manning, professor\_of, Stanford>} is extracted from evidence. Therefore, with the universal schema model, \textit{<Manning, member\_of, Stanford>} will be assigned a high probability given \textit{<Manning, prof\_of, Stanford>} is observed as evidence, and the gap is filled.

We evaluate our approach on the FEVER~\citep{thorne2018fever}, FEVER-Symmetric~\citep{schuster-etal-2019-towards}, FEVER 2.0~\citep{thorne-etal-2019-fever2}, and Climate-FEVER~\citep{diggelmann2020climate} datasets. Our findings  show that our system consistently outperforms zero-shot NLI model baselines by a margin of approximately 2.5 percentage points and beats the previous zero-shot approach by around 3 percentage points on FEVER-Symmetric. Notably, in contrast to state-of-the-art supervised methods~\cite{dehaven2023bevers}, our approach exhibits robustness on both adversarial datasets. When evaluated on the out-of-domain Climate-FEVER dataset, our method outperforms the supervised method by a margin exceeding 10 percentage points. 

\section{Related Work}
Recent advances in natural language processing have highlighted significant challenges associated with supervised learning models. A prominent concern is the models' tendency to learn dataset-specific biases, often at the expense of genuine linguistic understanding. 
For instance, ~\citet{schuster-etal-2019-towards} demonstrated the effectiveness of a claim-only model that classifies each claim in isolation, without the need for associated evidence. The high performance achieved by their system over the baseline can be attributed to the idiosyncrasies inherent in the dataset's construction. Similarly, ~\citet{thorne-etal-2019-fever2} highlighted the vulnerability of several FEVER systems, observing significant performance declines under adversarial conditions with simple rule-based perturbations. 
In other tasks such as NLI, previous works~\citep{poliak-etal-2018-hypothesis, gururangan-etal-2018-annotation} examined the susceptibility of neural models to such spurious correlations, revealing a troubling propensity for models to exploit unintended, data-specific heuristics. Taken together, these findings suggest that annotation artifacts within datasets contain discernible patterns. Such vulnerabilities underscore the necessity for more rigorous evaluation mechanisms, thus motivating the introduction of several adversarial fact-checking evaluation datasets~\citep{guo2022survey}.

\citet{pan-etal-2021-zero} presented the first work to investigate zero-shot fact verification, where they proposed a framework named Question Answering for Claim Generation (QACG). From any given evidence, QACG generates \textsc{Supports}, \textsc{Refutes}, and \textsc{NEI} claims. A classifier
is then trained using the generated claims instead of annotated claims, hence a zero-shot setting. To generate claims, QACG first produces QA pairs using a Question Generator fine-tuned on the processed SQuAD dataset~\citep{zhou2018neural}. Next, a QA-to-Claim Model is fine-tuned on the QA2D dataset~\citep{demszky2018transforming}, which converts each QA pair into a declarative sentence. However, their experiments are limited, using only the gold evidence to evaluate various zero-shot methods, which is not practical in a real-world setting. Also, unlike their work, where training is still performed using the generated training data, our approach does not require any training for claim verification.

Knowledge graphs have long been investigated in NLP, where the first discussions of a graphical knowledge representation can date back to the 50s~\citep{Newell1959ReportOA}. Since then, many NLP researchers have tried to integrate knowledge graphs into various NLP tasks, notably language models with knowledge graphs~\citep{nakashole-mitchell-2014-language, logan2019barack, liu2020k, wang2021kepler} and many downstream tasks such as question answering~\citep{liu2020k} and text classification~\citep{hu2021knowledgeable}. For fact-checking specifically,~\citet{ciampaglia2015computational}
proposed to use knowledge graphs to verify simple natural language claims, considering fact-checking as a special case of link prediction. Their method uses the subject and object of the claim and then finds the shortest path between the two entities. If the claim is true, there should be such a shortest path (or an edge); otherwise, there should be no shortest path (nor edge). While the fact that a simple shortest-path computation can assess the truth of new claims is exciting, this work is limited because all the factual claims are automatically generated using triples. Therefore, it does not directly apply to recent human-generated fact-checking datasets such as FEVER, as claims in FEVER are much more complicated. 

\section{Methodology} \label{Claim Verification}

As introduced, the verdict prediction step of claim verification is to predict a label $\mathcal{Y} \in \{\textsc{Supports}, \textsc{Refutes}, \textsc{NEI}\}$ given a claim $\mathcal{C}$ and its corresponding evidence $\mathcal{E}$, indicating if $\mathcal{C}$ is supported, refuted, or cannot be verified by $\mathcal{E}$. While we do not use any training data (manually or automatically labeled),
we assume a human-annotated development set is available for fine-tuning hyperparameters of our system. In keeping with prior research, we use the same set of notations and extend it to include triples. Table \ref{table:notations} contains all the notations used in the methodology. 

Figure \ref{fig:overview} illustrates the structure of our system, which comprises three main steps: Triple Extraction, Triple-level Verification, and Claim-level Verification. Additionally, we have integrated an external component, the Universal Schema. This section provides comprehensive insights into each component, outlining their functionalities and operations.

\begin{table}[]
    \begin{tabular}{c|l}
    notation & description \\ \hline
    $\mathcal{C}$     &       claim      \\ 
    $\mathcal{E}$     &      evidence \\ 
    $\mathcal{Y}$     &      claim-level label of $\mathcal{C}$  \\
    $\mathbb{C}$     &      set of triples extracted from $\mathcal{C}$ \\  
    $\mathbb{E}$     &      set of triples extracted from $\mathcal{E}$ \\  
    $\mathsf{c}$     &      $\mathsf{c} \in \mathbb{C}$ \\  
    $\mathsf{e}$     &      $\mathsf{e} \in \mathbb{E}$ \\  
    $\mathsf{y_e}$   &      triple-level label of $\mathsf{c}$ predicted by $\mathsf{e}$\\
    $\mathsf{y}$     &      aggregated triple-level label of $\mathsf{c}$ \\
    \end{tabular}
    \caption{Notations used in our fact-checking system}
    \label{table:notations}
\end{table}

\subsection{Triple Extraction}
A semantic triple consists of three entities: the subject, the object, and the relation between them. We denote such a triple as \textit{<subj, rel, obj>} where all three entities are natural language words, phrases, or clauses, and no schema needs to be specified in advance. Extracting a set of triples from plain text is called open information extraction (Open IE)~\citep{yates-etal-2007-textrunner}. 

As illustrated in Figure \ref{fig:overview}, our system first employs an OpenIE tool to extract triples from claim $\mathcal{C}$ and evidence $\mathcal{E}$, resulting in a set $\mathbb{C}$ of claim triples and a set $\mathbb{E}$ of evidence triples. Note that this step is back-traceable. For example, for any evidence triple, we can trace back which evidence sentence it comes from and which part of that sentence forms such evidence triple.

\subsection{Triple-level Verification} \label{Triple-level Verification} 
Given a claim triple $\mathsf{c}$ and a set of evidence triples $\mathbb{E}$, triple-level verification predicts a label $\mathsf{y} \in \{\textsc{Supports}, \textsc{Refutes}, \textsc{NEI}\}$ for this triple. Intuitively, any evidence triple may provide signals to predicting $\mathsf{y}$. Therefore, given $\mathsf{c}$, for each evidence triple $\mathsf{e}$ in $\mathbb{E}$, we utilise an NLI model to predict a label $\mathsf{y_p}$ with a softmax score as its probability. We linearise $\mathsf{c}$ and $\mathsf{e}$ and concatenate them with a special separation token: $\mathsf{e}$ [SEP] $\mathsf{c}$. For example, let $\mathsf{c} =$ \textit{<Barack Obama, was born in, USA>} and $\mathsf{e} =$ \textit{<Barack Obama, was born in, Hawaii>}, the input of the NLI model therefore is, `Barack Obama was born in Hawaii [SEP] Barack Obama was born in USA'. Following previous work, we map NLP labels to fact-checking labels, namely \textsc{Entailment} to \textsc{Supports}, \textsc{Contradiction} to \textsc{Refutes}, and \textsc{Neutral} to \textsc{NEI}.

To filter out less reliable triple-level labels $\mathsf{y_p}$, we set up two thresholds for \textsc{Supports} and \textsc{Refutes} as hyperparameters to cut off labels with low probabilities. The remaining labels are aggregated to reach a triple-level label $\mathsf{y}$ for the claim triple $\mathsf{c}$ using one of the following voting mechanisms:

\textbf{Max voting} takes the label with the overall highest probability as the triple-level label.

\textbf{Majority voting} ensures that the label with the most supporters (i.e., most frequent appearances) is the triple-level label. 

\textbf{Weighted sampling} samples a label according to the highest probabilities of each label.

Note that if all labels are filtered out, the triple-level label is \textsc{NEI} because none of the evidence triple is reliable enough for this claim triple.

\subsection{Claim-level Verification} \label{Claim-level Verification}
For each claim triple $\mathsf{c}$ in $\mathbb{C}$, a triple-level label $\mathsf{y}$ is predicted by the previous step. The final step is to reach a claim-level label $\mathcal{Y}$ from this set of triple-level labels using the following rule-based system also used by previous research~\citep{stacey-etal-2022-logical}: 

\begin{short_list}
If there exists a $\mathsf{y}$ that is \textsc{Refutes}, then $\mathcal{Y}$ is \textsc{Refutes}.
If no $\mathsf{y}$ is \textsc{Refutes} and there exists a $\mathsf{y}$ that is \textsc{NEI}, then $\mathcal{Y}$ is \textsc{NEI}.
Otherwise, $\mathcal{Y}$ is \textsc{Supports}.
\end{short_list}

\subsection{Universal Schema}
The challenge of integrating a  knowledge graph with our system stems from the incompatibility between a pre-determined schema and the unrestricted textual information extracted from open sources. In response, we put forward a solution that involves the implementation of the universal schema~\citep{riedel-etal-2013-relation}, which acts as an interface between pre-defined symbolic relations such as those found in knowledge graphs, and unconstrained textual relations such as those extracted by Open IE. Universal schema can be viewed as a  matrix that represents the knowledge base, comprising pairs of entities and relations. 

Notably, the original knowledge graph dataset employed in previous research on Universal Schema, namely Freebase~\citep{freebase}, is no longer maintained. Therefore, we undertook the task of training a novel universal schema model, utilizing a more contemporary language model architecture and incorporating data from Wikidata~\citep{wikidata} and some corresponding texts from Wikipedia. 

\rparagraph{Task Definition}
Consistent with ~\citet{riedel-etal-2013-relation}, a fact, or relation instance, is denoted by the pair \textit{rel} and \textit{<subj, obj>}. The goal of a universal schema model is, by definition, to estimate, for a given relation \textit{rel} and a given tuple \textit{<subj, obj>}, the probability $p (y_{rel,<subj, obj>} = 1)$ where the random variable $y_{rel,<subj, obj>}$ represents if \textit{<subj, obj>} is in relation \textit{rel}. In the context of our fact-checking scenario, we leverage these \textit{<subj, rel, obj>} triples to complete missing information.

\rparagraph{Objective}
To train our model, we adopt Bayesian Personalized Ranking (BPR)~\citep{bpr}. In this approach, observed true facts are assigned higher scores compared to both true and false unobserved facts. This scoring scheme serves as our optimization objective. Let $\sigma$ denote the sigmoid function, $\theta_{f+}$ denote the dot product of the latent representations of a positive ($\theta_{f-}$ for negative) fact pair \textit{rel} and \textit{<subj, obj>}, then the objective function is Obj$_{f+, f-} = -log(\sigma (\theta_{f+} - \theta_{f-}))$.

\rparagraph{Integration}
Upon successful training of the universal schema model, it becomes feasible to predict the probability of a tuple being associated with a given relation. The integration of this component into our fact-checking system involves utilizing the universal schema model to assign scores to potential triple candidates for a given set of claim triples $\mathbb{C}$ and supporting evidence triple set $\mathbb{E}$. All possible combinations of relations in the $\mathbb{C}$ and tuples in the $\mathbb{E}$ are considered as triple candidates. The universal schema model is then used to compute the probability of each triple candidate being true. Similar to Section \ref{Triple-level Verification}, a threshold is set to remove less reliable triple candidates. The triple candidates with a probability above the threshold are only utilized for triple-level verification if the available evidence triples are insufficient, i.e.\ when the label for the triple-level label $\mathsf{y}$ is \textsc{NEI}. In a manner akin to in-context learning, we also modify the Universal Schema model during the inference stage, upon encountering newly observed facts derived from evidence triples $\mathbb{E}$.

\section{Implementation}

\rparagraph{Evidence Retrieval}
To perform document level retrieval, we adopt the approach proposed by~\citet{hanselowski-etal-2018-ukp} For sentence level retrieval, we aim to demonstrate the effectiveness of our verification system without relying on any fact-checking training data. Therefore, we utilize traditional information retrieval techniques such as tf-idf weighting. In addition, we incorporate a semantic score as a weight factor, which is computed using the cosine similarity of embeddings generated by a neural model called stsb-roberta-base~\citep{reimers-gurevych-2019-sentence}.\footnote{https://huggingface.co/sentence-transformers/stsb-roberta-base}

\rparagraph{OpenIE Model}
We utilized an AllenNLP reimplementation of a BiLSTM sequence prediction model initially proposed by~\citet{stanovsky-etal-2018-supervised} as our Open Information Extraction (OpenIE) tool. The model can recognize verbs as relations and add their corresponding subjects and objects as arguments when given a sentence as input. For instances with more than two arguments, the model produces a triple for each combination of subjects or objects. If a relation only has one argument, known as a unary relation, a placeholder is added to ensure consistency across all generated triples.

\rparagraph{NLI Model}
In our experiments, we evaluate the effectiveness of our system using both base size and large size pre-trained NLI models. The aim is to demonstrate that our system consistently outperforms the NLI baselines. In particular, we leverage the RoBERTa base and large models, which have been pretrained on the MNLI dataset. Both models follow the standard NLI format of taking a premise and a hypothesis as input in the format of "[premise] SEP [hypothesis]", where SEP denotes the special separation token. We adhere to this format throughout our experiments.

\rparagraph{Universal Schema Model}
We leverage Sentence-BERT~\citep{reimers-gurevych-2019-sentence} to obtain sentence embeddings that serve as latent representations for both relations and tuples. This approach allows us to capture the semantic meaning of the sentences, which is essential for accurately representing the relations and tuples in our model. The pre-trained model "all-MiniLM-L6-v2"\footnote{https://huggingface.co/sentence-transformers/all-MiniLM-L6-v2} is utilized in our study, which is based on MiniLM~\citep{wang2020minilm}. This model has been pre-trained with a contrastive objective using diverse datasets containing sentence pairs. The cosine similarity is computed for each possible sentence pair within a batch, and cross-entropy loss is employed to compare these similarities with the true pairs.

\section{Experimental Setup}

\rparagraph{Dataset}
In our evaluation, we employ four benchmark datasets, FEVER, FEVER-Symmetric, FEVER 2.0, and Climate-FEVER. FEVER dataset~\cite{thorne2018fever} comprises 185,445 claims that are created by modifying sentences from Wikipedia, which are subsequently verified on Wikipedia without knowing the original sentence they were derived from. On the other hand, the FEVER-Symmetric dataset is introduced by~\citet{schuster-etal-2019-towards} to address the biases identified in the original FEVER dataset. This dataset is constructed with a regularization procedure to downweigh the giveaway phrases that cause potential biases. Similarly, the FEVER 2.0 dataset~\citep{thorne-etal-2019-fever2} comprises adversarial examples intentionally created by participants of the FEVER 2.0 shared task. The task required teams to generate claims specifically designed to challenge FEVER-trained models. From this dataset, we extracted all \textsc{Supports} and \textsc{Refutes} claims, along with their corresponding gold evidence sentences, for our evaluation. The Climate-FEVER dataset~\citep{diggelmann2020climate} is for verification of real-world climate change-related claims, excluding disputed claims.

To train our universal schema model, we utilize the Wikidata5m dataset~\citep{KEPLER}, which is a knowledge graph dataset comprising one million entities. This dataset is particularly suitable for our purposes, as it includes an aligned corpus, which we leverage in conjunction with the OpenIE tool to extract open-domain triples. In our approach, all triples from the Wikidata5m dataset and the extracted triples the aligned corpus are treated as positive samples. To generate negative samples based on a given positive sample, we utilize a randomized approach where we preserve the relation and generate arbitrary tuples that exist within the dataset. This approach allows us to create negative samples that differ from the positive samples while still being relevant to the original relation.

\rparagraph{Hyperparameters}
In our experiments, claim verification does not require model-level hyperparameter tuning since no training is involved. However, as outlined before, we have a small set of three thresholds to be adjusted: a threshold for the \textsc{Supports} label at the triple level, a threshold for the \textsc{Refutes} label at the triple level, and a threshold for filtering out Universal Schema triple candidates.\footnote{The specifics concerning the hyperparameters of the Universal Schema model can be found in Appendix \ref{app:hyper}. Note that the aforementioned thresholds were identified by conducting a search with a fixed Universal Schema model. } We adjusted their values on the FEVER dataset and did not perform any further adjustments on FEVER-Symmetric, FEVER 2.0, or Climate-FEVER. This deliberate choice was made to test the robustness of our system in handling different datasets without relying on dataset-specific optimization, an advantage of zero-shot approaches. 

\begin{figure}[]
    \centering
    \includegraphics[width=\columnwidth]{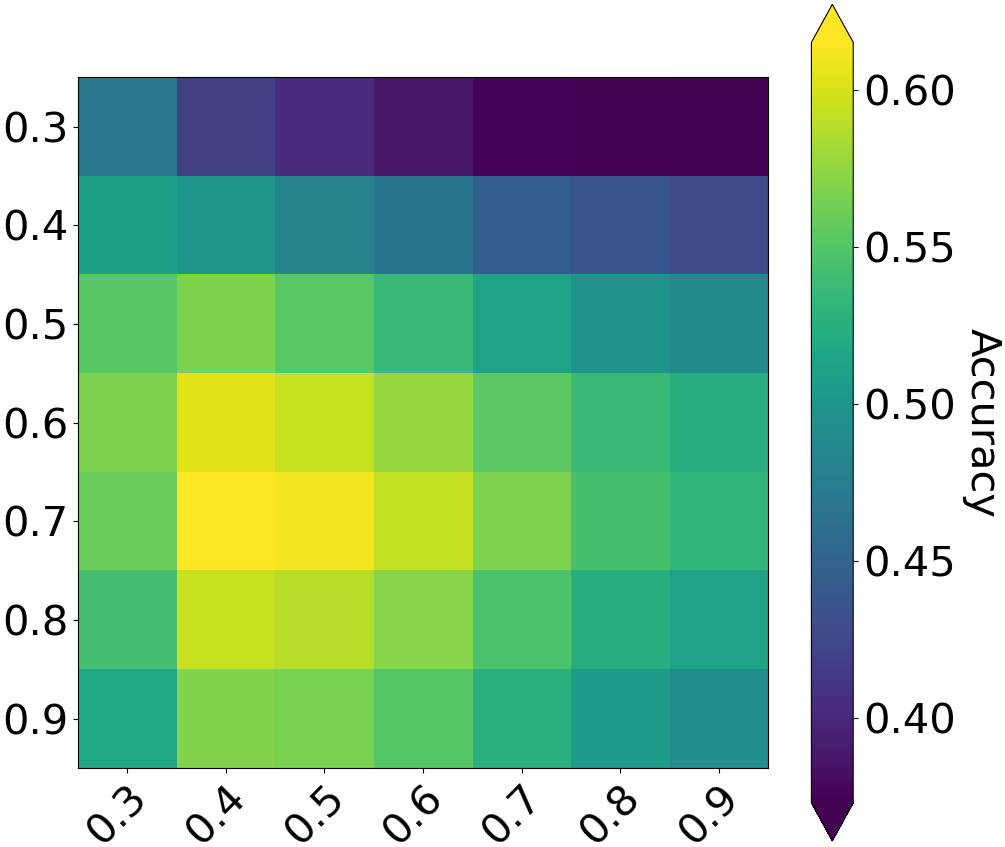}
    \caption{The influence of thresholds on accuracy for the \textsc{Supports} ($\mathsf{x}$-axis) and \textsc{Refutes} ($\mathsf{y}$-axis) with fixed threshold for Universal Schema triple candidates.} 
    \vspace{-3mm}
    \label{fig:threshold}
\end{figure}

Figures \ref{fig:threshold} illustrates the impact of thresholds on our system, and that optimizing them is relatively straightforward as the best settings are clustered in the region.
It is worth noting that the optimal threshold for \textsc{Refutes} is considerably higher compared to \textsc{Supports}, indicating that our system is more stringent in assigning a triple-level \textsc{Refutes} label than \textsc{Supports}. This difference is justified by the fact that, as explained in Section \ref{Claim-level Verification}, a single refuted claim triple is sufficient to refute the entire claim, therefore it helps being cautious when assigning a \textsc{Refutes} label to claim triples.

\section{Results and Discussion}
Our main results are presented in Table \ref{table:mainresult}, where we compare our system's performance against the current state-of-the-art system on FEVER, BEVERS~\citep{dehaven2023bevers}, the FEVER-trained entailment-predictor~\citep{diggelmann2020climate}, and QACG~\citep{pan-etal-2021-zero}\footnote{We made efforts to establish contact with the authors of QACG; however, our attempts to elicit a response were unsuccessful. Therefore, a direct comparison with their approach is not feasible, except for the FEVER-Symmetric dataset, where they reported performance under the same setting as ours.}. Also, we conduct additional ablation experiments to demonstrate the robustness of our system by varying the weighting factor and voting mechanism.

\begin{table*}[!t]
\begin{center}
\resizebox{\textwidth}{!}{
\begin{tabular}{lccccccccc}
\toprule
\multirow{2}{*}{\textbf{Model}} & \multicolumn{2}{c}{\textbf{FEVER}}                                                 &  & \multicolumn{1}{c}{\textbf{FEVER-Symmetric}} &  & \multicolumn{1}{c}{\textbf{FEVER 2.0}} &  & \multicolumn{2}{c}{\textbf{Climate-FEVER}}                                                           \\ \cmidrule(lr){2-3} \cmidrule(lr){5-5} \cmidrule(lr){7-7} \cmidrule(lr){9-10}
                                              & Accuracy               & FEVER Score                                  &  & Accuracy &  & Accuracy &  & Accuracy & F1  \\ \cmidrule(lr){2-3} \cmidrule(lr){5-5} \cmidrule(lr){7-7} \cmidrule(lr){9-10} \textit{Supervised} \\                                              \\
BEVERS                                      & \textbf{80.24}                    & \textbf{77.70}                                          &  & 75.9 & & 63.4 & & - & -                                          \\
\citet{diggelmann2020climate}  & 77.69                    & -                                          &  & -  &  & - & & 38.78 & 32.85                                          \\
\cmidrule(lr){2-3} \cmidrule(lr){5-5} \cmidrule(lr){7-7} \cmidrule(lr){9-10} \textit{Zero-shot} & \\
Random Guess                        & 33.33 & -  &  & 50.00 & & 50.00  & & 33.33 & 33.33   \\
QACG                        & - & -  &  & 77.1 & & - & & - & - \\
NLI Model (base)                             & 36.07 & 31.65 &  & 51.74 & & 49.68  & & - & - \\
NLI Model (large)                             & 58.12 & 54.38 &  & 78.94 & & 70.38  & & 44.37 & 44.24\\
Our system (base)                           & 38.56 & 33.73 &  & 56.62 & & 53.28 & & - & - \\
Our system (large)                           & 60.40 & 56.79 &  & \textbf{79.78} & & 72.92 & & \textbf{46.71} & \textbf{45.71}  \\
Our system (large) + USchema                         & \textbf{61.30}  & \textbf{57.84} &  & \textbf{79.78} & & \textbf{73.34} & & \textbf{46.71} & \textbf{45.71}  \\\bottomrule
\end{tabular}
}
\end{center}
\caption{Main results on FEVER (S/R/NEI), FEVER-Symmetric (S/R), FEVER 2.0 (S/R) and Climate-FEVER (S/R/NEI). BEVERS is the current state-of-the-art system on FEVER and \citet{diggelmann2020climate} is the entailment-predictor based on ALBERT (large-v2). The accuracy on FEVER-Symmetric, FEVER 2.0 and Climate-FEVER datasets was achieved without fine-tuning, demonstrating the models' robustness.}
\label{table:mainresult}
\end{table*}

\begin{table}[htp]
\def\arraystretch{0.999}
\begin{center}
\resizebox{.4\textwidth}{!}
{
    \begin{tabular}{lc}
    \toprule
    \multirow{1}{*}{\bf Variant} & \multicolumn{1}{c}{\bf $\Delta$ Accuracy}  \\ \cmidrule(lr){2-2}
    tf-idf                            & +\textbf{2.28}                     \\ 
    Cosine similarity               & +2.11 \\ \cmidrule(lr){1-2}
    Max voting                    & +2.05 \\
    Majority voting                  & +2.13 \\
    Weighted sampling                          & +1.93 \\ \bottomrule
    \end{tabular} 
}
\caption{Improvements of our system over baseline using different retrieval weighting factor and voting technique are steady.}

\vspace{-5mm}

\label{tab:delta1}
\end{center}
\end{table}

\begin{finding}
Our zero-shot approach exhibited enhanced robustness against adversarial perturbations and manifested notable out-of-domain effectiveness in contrast to supervised approaches.
\end{finding}
As shown in Table \ref{table:mainresult}, our zero-shot method demonstrates greater resilience against adversarial attacks compared to supervised methods, providing a significant advantage in real-world scenarios where the presence of misinformation and deceptive tactics can impede the performance of fact-checking systems. 

By abstaining from using training data, our approach intuitively circumvents these issues and offers a more robust approach to fact-checking. Specifically, despite exhibiting lower performance than supervised systems on the original FEVER dataset, our models achieved highly competitive scores on the FEVER-Symmetric dataset, trailing the state-of-the-art by only approximately 2 percentage points. We attribute this positive outcome to our system's utilization of NLI models, which already demonstrate outstanding performance on this adversarial dataset. The results obtained on the FEVER 2.0 dataset align with FEVER-Symmetric and further strengthen our conclusions.

On Climate-FEVER, the supervised approach delineated by ~\citet{diggelmann2020climate} achieved an accuracy of 38.78\% and an F1 score of 32.85\%. In comparison, our introduced zero-shot methodology showcased enhanced results, achieving an accuracy rate of 46.71\% and an F1 score of 45.71\%, which demonstrated a notable generalization ability. These findings suggest that our zero-shot method offers a promising avenue for improved performance in out-of-domain tasks.

\begin{table}[htp]
\def\arraystretch{0.999}
\begin{center}
\resizebox{.4\textwidth}{!}
{
\begin{tabular}{lc}
\toprule
\multirow{1}{*}{\bf Evidence} & \multicolumn{1}{c}{\bf $\Delta$ Accuracy}  \\ \cmidrule(lr){2-2}
Gold + Random                          & +\textbf{7.93}                     \\ 
Gold + Retrieved (tf-idf)              & +3.74 \\
Retrieved (tf-idf)                         & +2.28 \\ \bottomrule
\end{tabular}
}
\caption{Improvements of our system over baseline using gold evidence vs. retrieved evidence. }

\vspace{-5mm}

\label{tab:delta2}
\end{center}
\end{table}

\begin{table*}[!t]
\resizebox{\textwidth}{!}{
\begin{tabular}{|l|l|}
\hline

\textbf{Claim} & \makecell{The Adventures of Pluto Nash was \textcolor{red}{reviewed} by Ron Underwood .} \\ \hline

\textbf{Evidence Sentences} & \makecell{1: The Adventures of Pluto Nash is a 2002 Australian-American \\  science fiction action comedy film starring Eddie Murphy \\ -LRB- in a dual role -RRB- and directed by Ron Underwood . \\ 2: Ron `` Thunderwood '' Underwood is a musician and director from Phoenix , Arizona .\\ ...} \\ \hline

\textbf{Evidence Triples} & \makecell{<a 2002 Australian - American science fiction action comedy film, starring, Eddie Murphy>  \\ ...} \\ \hline

\textbf{USchema Triples} & \makecell{<The Adventures of Pluto Nash, \textcolor{red}{directed},  by Ron Underwood> \\ ...} \\ \hline

\end{tabular}
}
\caption{An example with Universal Schema triples. Due to space limitations, not all sentences and triples for this example are shown. The table focuses on the most critical ones that effectively demonstrate our points.}
\label{table:examples}
\end{table*}

\begin{finding}
Our system, utilizing triple-level inference, consistently improves over the baseline results irrespective of the NLI model used.
\end{finding}
In our experiments, our approach was able to improve the performance of both the base size and large size NLI models by approximately 2.5\%. These consistent improvements suggest that our approach can continue to benefit from the ongoing progress: as more advanced models are being developed, our system is expected to demonstrate even greater accuracy and reliability.

In addition, we performed ablation experiments to investigate the impact of various weighting factors and different voting mechanisms, as outlined in Section \ref{Claim Verification}. The results, presented in Table \ref{tab:delta1}, demonstrate that our system's improvements over the baseline NLI models in Table \ref{table:mainresult} are consistently observed across all variants, indicating the reliability and robustness of our approach.

Furthermore, we conducted experiments to evaluate the effect of evidence quality on claim verification, as presented in Table \ref{tab:delta2}. The Gold + Random method involves using gold-standard evidence for \textsc{Supports} and \textsc{Refutes} claims, while random evidence is used for \textsc{NEI} claims. The Gold + Retrieved method is similar, but uses retrieved evidence instead of random evidence for \textsc{NEI} claims while still utilizing gold-standard evidence for \textsc{Supports} and \textsc{Refutes} claims. The results indicate that the performance improvements of our system increases as the quality of evidence improves, suggesting that our zero-shot approach benefits more from less noisy evidence. This likely due to the fact that our system relies on a strict set of rules to classify claims, which may be more sensitive to the presence of noise in the evidence. Thus, our system is likely to benefit from the continued development of better evidence retrieval systems.

\begin{finding}
Employing the Universal Schema model provides marginal gains by bridging the gaps between extracted claim and evidence triples.
\end{finding}
The Universal Schema model, despite its modest gains, contributes to enhancing the overall performance of our fact-checking system. In manual analysis of the results we found that integrating the Universal Schema model helps our approach in handling claims involving mutual exclusivity, resulting in increased accuracy. Mutual exclusivity denotes a situation in which two or more events cannot coexist simultaneously. To illustrate this, let us consider the claim in Table \ref{table:examples} \textit{The Adventures of Pluto Nash was reviewed by Ron Underwood}, initially classified as \textsc{Not Enough Information} (\textsc{NEI}) in the absence of the Universal Schema model. This misclassification originated from the complexity of the retrieved evidence, which presented a complex sentence implying that Ron Underwood directed the movie, thereby refuting the claim. However, extracting the relation needed as evidence \textit{<The Adventures of Pluto Nash, directed by, Ron Underwood>} posed challenges so it was not extracted. Consequently, due to the absence of this critical triple, the model erroneously labeled the claim as \textsc{NEI}. By incorporating the Universal Schema model, our system successfully recovered the missing evidence triple, while also recognizing the inherent mutual exclusivity between assuming both the director and reviewer roles for the same movie. As a result, using the Universal Schema model accurately predicted the \textsc{Refutes} label.

We also observed that the Universal Schema model offers limited assistance when applied to the two adversarial datasets considered. This is due to the fact that, in both the FEVER-Symmetric and FEVER 2.0 setting, all the necessary evidence is provided, unlike real-world scenarios. Consequently, the value provided by the Universal Schema model, which primarily focuses on filling gaps, becomes minimal since no gaps exist in the presence of sufficient evidence.

\section{Conclusion}
We introduced a novel zero-shot fact-checking method, translating claims and evidence into semantic triples with external knowledge graphs. This method surpasses other zero-shot baselines, impressively without direct FEVER dataset training. Its resilience is evident, avoiding the typical performance dips seen in supervised models on adversarial datasets like FEVER-Symmetric and FEVER 2.0. Also on the Climate-FEVER dataset, our approach outshines even supervised counterparts, highlighting its generalization prowess. Augmented by pretrained NLI models, our system's robustness is further emphasized. As future steps, we aim to hone model interpretability, examine diverse knowledge graphs, and test our method's versatility on other fact-checking datasets.

\section*{Limitations}
While our novel zero-shot learning method for fact-checking with semantic triples and knowledge graphs has shown promising results, there are several limitations that must be noted.

Firstly, our method's language capabilities have been exclusively tested on the English language, which poses an inherent limitation. Though the method was not specifically designed and implemented for English, the experiments were solely conducted using English datasets. Consequently, the potential effectiveness of our approach with other languages remains unverified. Differences in linguistic features and semantic triple structures across languages might present unique challenges that we have yet to encounter or address.

Secondly, our approach relies heavily on Wikipedia as both the source for datasets used in evaluation and the basis for our knowledge graphs. While Wikipedia is a vast and continually updated source of knowledge, its use as the sole source of data introduces biases and limitations. Wikipedia's content is predominantly generated by its user community, which can lead to the inclusion of inaccuracies, cultural biases, or omissions. This limitation might affect the fact-checking capabilities of our model, as the reliability of its responses are directly proportional to the quality and accuracy of the information within Wikipedia.

Additionally, the reliance on a single source for data and knowledge graphs constrains the method's applicability in fact-checking scenarios where knowledge outside of Wikipedia's domain is required. It may also lead to an overfitting issue, as the model might be overly tuned to Wikipedia's style and structure, limiting its performance when applied to different or broader sources.

In future work, addressing these limitations by incorporating support for multiple languages and expanding the data sources beyond Wikipedia would be essential steps towards enhancing the effectiveness and generalizability of our approach.

\section*{Ethics Statement}
The use of fact-checking datasets and systems has become increasingly important in combatting misinformation, and as such, it is necessary to consider the ethical implications of their use. One of the key concerns in this regard is the potential for biases in these datasets. Such biases can arise from various sources, including the selection and interpretation of sources, the types of claims being fact-checked, and the demographic characteristics of the individuals involved. These biases have the potential to perpetuate stereotypes and reinforce existing power dynamics, and thus it is the responsibility of researchers to ensure that they use representative and unbiased datasets to train and evaluate their models. Transparency regarding any potential biases in models is also essential, and steps must be taken to mitigate any negative impact. By addressing these ethical concerns, researchers can promote the integrity of fact-checking and contribute to a more informed and equitable public discourse.

\section*{Acknowledgements}
Zhangdie Yuan and Andreas Vlachos are both supported by the ERC grant AVeriTeC (GA 865958).

\bibliography{anthology,custom}

\appendix
\section{Hyperparameters}\label{app:hyper}
For the universal schema model, the hyperparameters were manually tuned using the wikidata dev set, with a batch size of 32, a learning rate of 2e-5, an Adam epsilon of 1e-8, a weighted decay of 0.01 and a maximum gradient norm of 1.0. The model was trained for a maximum of 3 epochs, with early stopping based on the loss observed on the dev set. Given the large amount of training data and limited computing resources, we partition the data into sections of 10000000 randomly shuffled samples to make the task feasible. Each section is treated as a separate training batch for our model.

\end{document}